\documentclass{ieeeaccess}
\usepackage{cite}
\usepackage{amsmath,amssymb,amsfonts}
\usepackage{graphicx}
\graphicspath{{imgs/}}
\usepackage[labelformat=simple]{subfig}
\usepackage{textcomp}

\usepackage{pifont}
\newcommand{\cmark}{\ding{51}}%
\newcommand{\xmark}{\ding{55}}%

\begin{document}
\history{Date of publication xxxx 00, 0000, date of current version xxxx 00, 0000.}
\doi{10.1109/ACCESS.2020.2980266}

\title{Convolutional Neural Networks with Intermediate Loss for 3D Super-Resolution of CT and MRI Scans}
\author{\uppercase{Mariana-Iuliana Georgescu}\authorrefmark{1}, \uppercase{{Radu Tudor} Ionescu\authorrefmark{1},
\IEEEmembership{Member, IEEE}, and Nicolae Verga}\authorrefmark{2,3}}
\address[1]{University of Bucharest, Bucharest, Romania}
\address[2]{``Carol Davila'' University of Medicine and Pharmacy, Bucharest, Romania}
\address[3]{Colțea Hospital, Bucharest, Romania}
\tfootnote{The research leading to these results has received funding from the EEA Grants 2014-2021, under Project contract no. EEA-RO-NO-2018-0496.}

\markboth
{Georgescu \headeretal: Convolutional Neural Networks With Intermediate Loss for 3D Super-Resolution of CT and MRI Scans}
{Georgescu \headeretal: Convolutional Neural Networks With Intermediate Loss for 3D Super-Resolution of CT and MRI Scans}

\corresp{Corresponding author: Radu Tudor Ionescu (e-mail: raducu.ionescu@gmail.com).}

\begin{abstract}
Computed Tomography (CT) scanners that are commonly-used in hospitals and medical centers nowadays produce low-resolution images, e.g. one voxel in the image corresponds to at most one-cubic millimeter of tissue. In order to accurately segment tumors and make treatment plans, radiologists and oncologists need CT scans of higher resolution. The same problem appears in Magnetic Resonance Imaging (MRI). In this paper, we propose an approach for the single-image super-resolution of 3D CT or MRI scans. Our method is based on deep convolutional neural networks (CNNs) composed of 10 convolutional layers and an intermediate upscaling layer that is placed after the first 6 convolutional layers. Our first CNN, which increases the resolution on two axes (width and height), is followed by a second CNN, which increases the resolution on the third axis (depth). Different from other methods, we compute the loss with respect to the ground-truth high-resolution image right after the upscaling layer, in addition to computing the loss after the last convolutional layer. The intermediate loss forces our network to produce a better output, closer to the ground-truth. A widely-used approach to obtain sharp results is to add Gaussian blur using a fixed standard deviation. In order to avoid overfitting to a fixed standard deviation, we apply Gaussian smoothing with various standard deviations, unlike other approaches. We evaluate the proposed method in the context of 2D and 3D super-resolution of CT and MRI scans from two databases, comparing it to related works from the literature and baselines based on various interpolation schemes, using $2\times$ and $4\times$ scaling factors. The empirical study shows that our approach attains superior results to all other methods. Moreover, our subjective image quality assessment by human observers reveals that both doctors and regular annotators chose our method in favor of Lanczos interpolation in $97.55\%$ cases for an upscaling factor of $2\times$ and in $96.69\%$ cases for an upscaling factor of $4\times$. In order to allow others to reproduce our state-of-the-art results, we provide our code as open source at {https://github.com/lilygeorgescu/3d-super-res-cnn}.
\end{abstract}


\titlepgskip=-15pt

\maketitle

\section{Introduction}

\begin{figure*}[!th]
\centering
\includegraphics[width=0.6\linewidth]{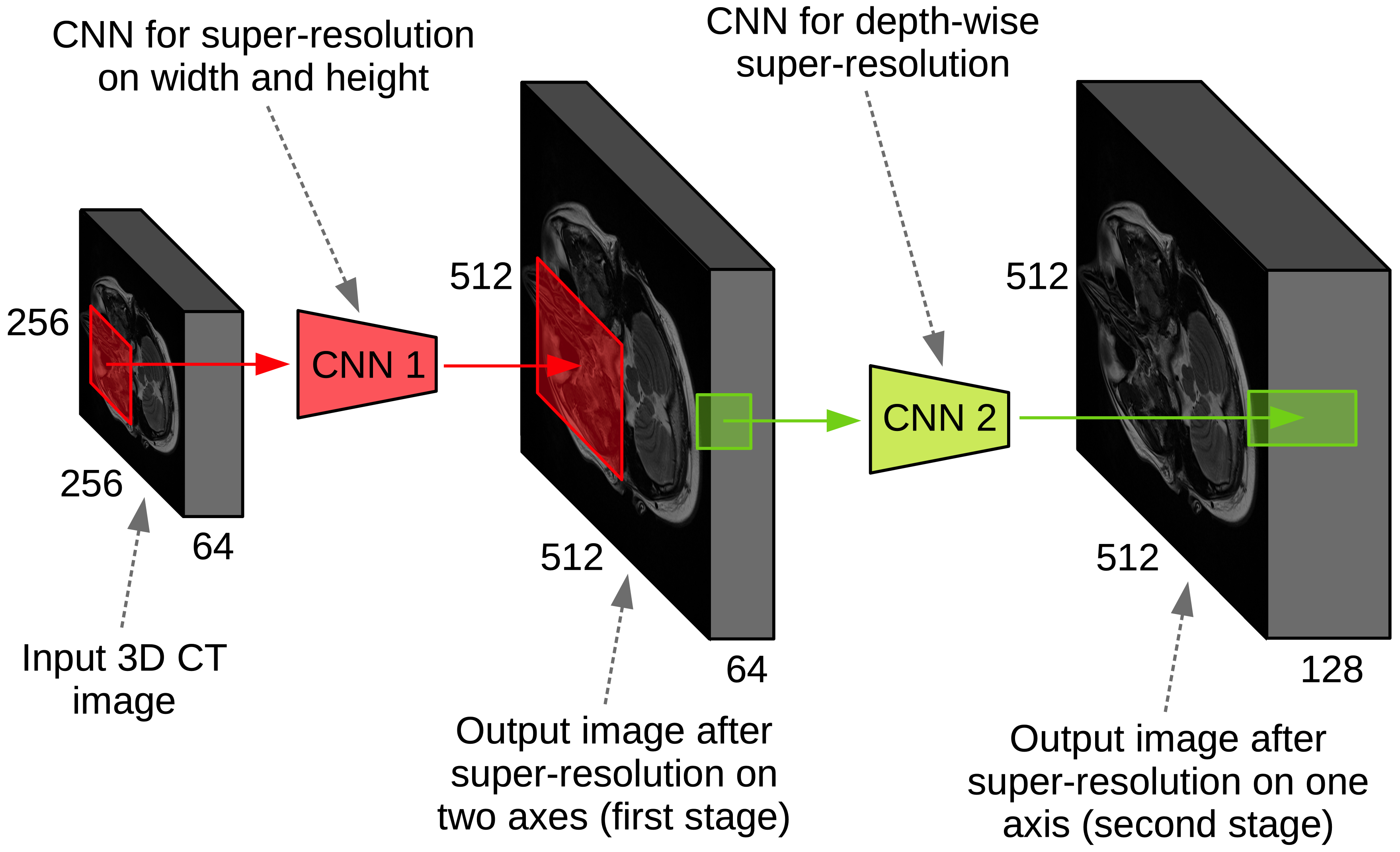}
\caption{Our method for 3D image super-resolution based on two subsequent fully convolutional neural networks. In the first stage, the input volume is resized in two dimensions (width and height). In the second stage, the processed volume is further resized in the third dimension (depth). Using a scale factor of $2\times$, an input volume of $256\times256\times64$ components is upsampled to $512\times512\times128$ components (on all axes). Best viewed in color.}
\label{fig_pipeline}
\end{figure*}

\IEEEPARstart{M}{edical} centers and hospitals around the globe are typically equipped with single-energy Computer Tomography (CT) or Magnetic Resonance Imaging (MRI) scanners that produce cross-sectional images (slices) of various body parts. The resulting images are of low-resolution, since one pixel usually corresponds to at most one-millimeter piece of tissue. The thickness of one slice is one millimeter at best, so the 3D CT images are composed of volumetric pixels (voxels) that usually correspond to one cubic millimeter ($1 \times 1 \times 1\;mm^3$) of tissue. One of the main benefits of this non-invasive scanning technique is that it allows doctors to see if there are malignant tumors inside the body. Nevertheless, doctors, and even machine learning systems \cite{Sert-MH-2019}, are not able to accurately contour (segment) the tumor regions because of the low-resolution of CT or MRI scans. According to a team of radiologists from Colțea Hospital in Bucharest, that provided a set of anonymized CT scans for our experiments, the desired resolution is to have one voxel correspond to one cubic micrometer (a thousandth part of a cubic millimeter) of tissue. In other words, the goal is to increase the resolution of 3D CT and MRI scans by a factor of $10\times$ in each direction.

The main motivation behind our work is to allow radiologists and oncologists to accurately segment tumors and make better treatment plans. In order to achieve the desired goal, we propose a machine learning method that takes as input a 3D image and increases the resolution of the input image by a factor of $2\times$ or $4\times$, providing as output a high-resolution 3D image. To our knowledge, there are only a few previous works \cite{Sert-MH-2019, Chen-MICCAI-2018, Du-AS-2019, Du-BIBM-2018, Du-NC-2019, Hatvani-TRPMS-2018, Hatvani-TMI-2018, Huang-CVPR-2017, Jurek-BBE-2019, Li-Access-2019, Mahapatra-CMIG-2019, Oktay-MICCAI-2016, Pham-CMIG-2019, Shi-JBHI-2018, You-TMI-2019, Yu-ICIP-2017, ZENG-CBM-2018, Zhao-TMI-2019} that study the super-resolution of CT or MRI images. Similar to some of these previous works \cite{Chen-MICCAI-2018, Du-AS-2019, Du-BIBM-2018, Du-NC-2019, Hatvani-TRPMS-2018, Hatvani-TMI-2018, Jurek-BBE-2019, Mahapatra-CMIG-2019, Oktay-MICCAI-2016, Pham-CMIG-2019, Shi-JBHI-2018, You-TMI-2019, Yu-ICIP-2017, ZENG-CBM-2018, Zhao-TMI-2019}, we approach single-image super-resolution (SISR) of CT and MRI scans using deep convolutional neural networks (CNNs). 
We propose a CNN architecture composed of 10 convolutional layers and an intermediate sub-pixel convolutional (upscaling) layer \cite{Shi-CVPR-2016} that is placed after the first 6 convolutional layers. Different from related works \cite{Du-AS-2019, Hatvani-TRPMS-2018, Yu-ICIP-2017, Zhao-TMI-2019} that use the sub-pixel convolutional layer of Shi et al.~\cite{Shi-CVPR-2016}, we add 4 convolutional layers after the upscaling layer. In order to obtain 3D super-resolution, we employ two CNNs with similar architectures, as illustrated in Figure~\ref{fig_pipeline}. The first CNN increases the resolution on two axes (width and height), while the second CNN takes the output from the first CNN and further increases the resolution on the third axis (depth). 
Different from related methods \cite{Du-AS-2019, Hatvani-TRPMS-2018, Yu-ICIP-2017, Zhao-TMI-2019}, we compute the loss with respect to the ground-truth high-resolution image right after the upscaling layer, in addition to computing the loss after the last convolutional layer. The intermediate loss forces our network to produce a better output, closer to the ground-truth. In order to improve the results and obtain sharper images, a common approach is to apply Gaussian smoothing on top of the input images, using a fixed standard deviation. Different from other medical image super-resolution methods \cite{Du-AS-2019, Shi-JBHI-2018, ZENG-CBM-2018}, we use various standard deviations in order to avoid overfitting to a certain standard deviation and improve the generalization capacity of our model.

We note that our model belongs to a class of deep neural networks known as fully convolutional neural networks. The main advantage of using such models, which do not include dense (fully-connected) layers, is that the input samples do not have to be of the same size. This flexibilty enables a broad range of applications such as image segmentation \cite{Long-CVPR-2015}, object tracking \cite{Wang-ICCV-2015}, crowd detection \cite{Castellano-SOFSEM-2020}, time series classification \cite{Karim-Access-2017} and single-image super-resolution \cite{Du-AS-2019, Hatvani-TRPMS-2018, Yu-ICIP-2017, Zhao-TMI-2019}. Different from other fully convolutional neural networks \cite{Du-AS-2019, Hatvani-TRPMS-2018, Yu-ICIP-2017, Zhao-TMI-2019,Long-CVPR-2015,Wang-ICCV-2015,Castellano-SOFSEM-2020,Karim-Access-2017}, our network is specifically designed for SISR, having a custom architecture that includes an upscaling layer \cite{Shi-CVPR-2016} useful only for SISR, as well as a novel loss function.

We conduct super-resolution experiments on two databases of 3D CT and MRI images, one gathered from the Colțea Hospital (CH) and one that is publicly available online, known as NAMIC\footnote{Available at {http://hdl.handle.net/1926/1687}.}. We compare our method with several interpolation baselines (nearest, bilinear, bicubic, Lanczos) and state-of-the-art methods \cite{Du-AS-2019, Pham-CMIG-2019, You-TMI-2019, ZENG-CBM-2018}, in terms of the peak signal-to-noise ratio (PSNR), the structural similarity index (SSIM) and the information fidelity criterion (IFC). We perform comparative experiments on both 2D and 3D single-image super-resolution under commonly-used upscaling factors, namely $2\times$ and $4\times$. The empirical results indicate that our approach is able to surpass all the other methods included in the experiments. For example, on the NAMIC data set, we obtain a PSNR of $40.57$ and an SSIM of $0.9835$ for 3D super-resolution by a factor of $2\times$, while Pham et al.~\cite{Pham-CMIG-2019} reported a PSNR of $38.28$ and an SSIM of $0.9781$ in the same setting. Furthermore, we conduct a subjective image quality assessment by human observers, asking 6 doctors and 12 regular annotators to choose between the CT images produced by our method and those produced by Lanczos interpolation (the best interpolation method). The annotators opted for our method in favor of Lanczos interpolation in $97.55\%$ cases for an upscaling factor of $2\times$ and in $96.69\%$ cases for an upscaling factor of $4\times$. These results indicate that our method is significantly better than Lanczos interpolation. In order to allow further developments and results replication, we provide our code as open source in a public repository\footnote{Available at {https://github.com/lilygeorgescu/3d-super-res-cnn}.}.

To summarize, our contribution is threefold:
\begin{itemize}
\item We propose a novel CNN model for 3D super-resolution of CT and MRI scans, which is based on an intermediate loss added to the standard output loss and on smoothing the input using random standard deviations for the Gaussian blur.
\item We conduct a subjective image quality assessment by human observers to determine the quality and the utility of our super-resolution results, as in \cite{You-TMI-2019}.
\item We provide our code online for download, allowing our results to be easily replicated.
\end{itemize}

We organize the rest of this paper as follows. We present related work in Section~\ref{sec_Related_Work}. We describe our method in detail in Section~\ref{sec_Method}. We present experiments and results in Section~\ref{sec_Experiments}. Finally, we draw our conclusions in Section~\ref{sec_Conclusion}.

\section{Related Work}
\label{sec_Related_Work}

The purpose of SISR is to reconstruct a high-resolution (HR) image from its low-resolution (LR) counterpart. Before the deep learning age, researchers have used exemplar-based or sparse coding methods for SISR. Exemplar-based methods learn mapping functions from external LR and HR exemplar pairs \cite{Bevilacqua-BMVC-2012, Chang-CVPR-2004, Dai-CGF-2015}. Sparse coding methods \cite{Yang-TIP-2010} are representative for external exemplar-based SR methods. For example, the method of Yang et al.~\cite{Yang-TIP-2010} builds a dictionary with LR patches and the corresponding HR patches. 

To our knowledge, the first work to present a deep learning approach for SISR is \cite{Dong-TPAMI-2016}. Dong et al.~\cite{Dong-TPAMI-2016} proposed a CNN composed of 8 convolutional layers. The network was trained in an end-to-end fashion, minimizing the reconstruction error between the HR image and the output of the network. They used bicubic interpolation to resize the image, before giving it as input to the network. Hence, the CNN takes a blurred HR image as input and learns how to make it sharper. Since the input is an HR image, this type of CNN is time consuming. Therefore, Shi et al.~\cite{Shi-CVPR-2016} introduced a new method for upsampling the image using the CNN activation maps produced by the last layer. Their network is more efficient, because it builds the HR image only at the very end. Other works, such as \cite{Zhang-ECCV-2018}, proposed deeper architectures, focusing strictly on accuracy. Indeed, Zhang et al.~\cite{Zhang-ECCV-2018} presented one of the deepest CNNs used for SR, composed of 400 layers. They used a channel attention mechanism and residual blocks to handle the depth of the network. 

For medical SISR, some researchers have focused on sparse representations \cite{Huang-CVPR-2017, Li-Access-2019}, while others on training convolutional neural networks \cite{Sert-MH-2019, Chen-MICCAI-2018, Du-AS-2019, Du-BIBM-2018, Du-NC-2019, Hatvani-TRPMS-2018, Hatvani-TMI-2018, Jurek-BBE-2019, Mahapatra-CMIG-2019, Oktay-MICCAI-2016, Shi-JBHI-2018, You-TMI-2019, Yu-ICIP-2017, Zhao-TMI-2019}. 

The authors of \cite{Huang-CVPR-2017} proposed a weakly-supervised joint convolutional sparse coding method to simultaneously solve the problems of super-resolution and cross-modal image synthesis. In \cite{Li-Access-2019}, the authors adopted a method based on compressed sensing and self-similarity constraint, obtaining better results than \cite{ZENG-CBM-2018} in terms of SSIM and PSNR.

Some works \cite{Sert-MH-2019, Du-AS-2019, Du-NC-2019, Hatvani-TRPMS-2018, Jurek-BBE-2019, Mahapatra-CMIG-2019, Li-Access-2019, Shi-JBHI-2018, You-TMI-2019, Yu-ICIP-2017, Zhao-TMI-2019} focused on 2D upsampling, i.e. on increasing the width and height of CT/MRI slices, while other works \cite{Chen-MICCAI-2018, Du-BIBM-2018, Huang-CVPR-2017, Oktay-MICCAI-2016} focused on 3D upsampling, i.e. on increasing the resolution of full 3D CT/MRI scans on all three axes (width, height and depth). 
  
For 2D upsampling, some works \cite{Sert-MH-2019, Du-NC-2019, Jurek-BBE-2019, Shi-JBHI-2018} used interpolated low resolution (ILR) images, while other works \cite{Du-AS-2019, Hatvani-TRPMS-2018, Yu-ICIP-2017, Zhao-TMI-2019} used the efficient sub-pixel convolutional neural network (ESPCN) introduced in \cite{Shi-CVPR-2016}. Similar to the latter approaches \cite{Du-AS-2019, Hatvani-TRPMS-2018, Yu-ICIP-2017, Zhao-TMI-2019}, we employ the sub-pixel convolutional layer of Shi et al.~\cite{Shi-CVPR-2016}. Different from these related works \cite{Du-AS-2019, Hatvani-TRPMS-2018, Yu-ICIP-2017, Zhao-TMI-2019}, we add a convolutional block after the sub-pixel convolutional layer, in order to enhance the HR output image. Furthermore, we propose a novel loss function for our CNN model. Instead of computing the loss between the output image and the ground-truth high-resolution image, we also compute the loss between the intermediate image given by the sub-pixel convolutional layer and the high-resolution image. This forces our neural network to learn a better intermediate representation, increasing its performance. 

There are some works \cite{Chen-MICCAI-2018, Mahapatra-CMIG-2019, You-TMI-2019} that employed generative adversarial networks (GANs) \cite{Goodfellow-NIPS-2014} to upsample medical images. Although our approach based on fully convolutional neural networks is less related to GAN-based SISR methods, we decided to include the approach of You et al.~\cite{You-TMI-2019} in our experiments, as a recent and relevant baseline.

For 3D upsampling, Chen et al.~\cite{Chen-MICCAI-2018} trained a CNN with 3D convolutions and used a GAN--based loss function to produce sharper and more realistic images. In order to upsample a 3D image, Du et al.~\cite{Du-BIBM-2018} employed a deconvolutional layer composed of 3D filters to upsample the LR image, in an attempt to reduce the computational complexity. As \cite{Chen-MICCAI-2018, Du-BIBM-2018, Huang-CVPR-2017, Oktay-MICCAI-2016}, we tackle the problem of 3D CT/MRI image super-resolution. However, instead of using inefficient 3D filters to upsample the LR images in a single step, we propose a more efficient two-stage approach that uses 2D filters. Our approach employs a CNN to increase the resolution in width and height, and another CNN to further increase the resolution depth-wise.

\begin{figure*}[t!]
\centering
\includegraphics[width=1.0\linewidth]{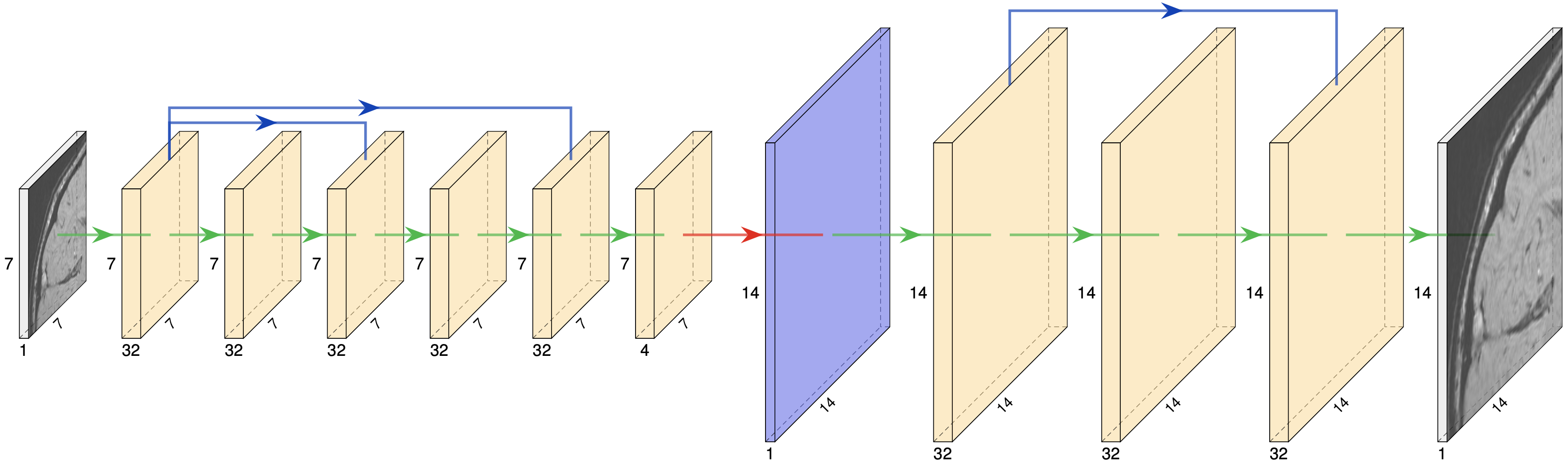}
\caption{Our convolutional neural network for super-resolution on two axes, height and width. The network is composed of 10 convolutional layers and an upsampling (sub-pixel convolutional) layer. It takes as input low-resolution patches of $7 \times 7$ pixels and, for the $r=2$ scale factor, it outputs high-resolution patches of $14 \times 14$ pixels. The convolutional layers are represented by green arrows. The sub-pixel convolutional layer is represented by the red arrow. The long-skip and short-skip connections are represented by blue arrows. Best viewed in color.}
\label{fig_architecture}
\end{figure*}

Most SISR works \cite{Du-AS-2019, Shi-JBHI-2018, ZENG-CBM-2018}, apply Gaussian smoothing using a fixed standard deviation on the training images, thus training the models in more difficult conditions. However, we believe that using a fixed standard deviation can harm the performance, as deep models tend to overfit to the training data. Different from the standard methodology, each time we apply smoothing on a training image, we chose a different standard deviation, randomly. This simple change improves the generalization capacity of our model, yielding better performance at test time.

While many works focus only on the super-resolution task, the work of Sert et al.~\cite{Sert-MH-2019} is focused on the gain brought by the upsampled images in solving a different task. Indeed, the authors \cite{Sert-MH-2019} obtained an improvement of $7.5\%$ in the classification of segmented brain tumors when the upsampled images were used.

We note that there is also some effort in designing and obtaining CT scan results of higher resolution directly from CT scanners. For example, X-ray microtomography (micro-CT) \cite{Elliott-JM-1982}, which is based on pixel sizes of the cross-sections in the micrometer range, has applications in medical imaging \cite{Enders-PO-2017, Tromba-AIP-2010}. Another alternative to standard (single-energy) CT is dual-energy or multi-energy CT \cite{McCollough-R-2015, Doerner-EJR-2017}. Different from the expensive alternatives such as dual-energy CT and micro-CT, our approach to increasing the resolution of single-energy CT images using a machine learning algorithm represents an economical and accessible mode.

\section{Method} 
\label{sec_Method}

Our method for solving the 3D image super-resolution problem relies on deep neural networks. The universal approximation theorem \cite{Cybenko-MCSS-1989,Hornik-NN-1991} states that neural networks with at least one hidden layer are universal function approximators. Hence, the hypothesis class represented by neural networks is large enough to accommodate any hypothesis explaining the data. This high model capacity seems to help deep neural networks in attaining state-of-the-art results in many domains \cite{LeCun-Nature-2015}, including image super-resolution \cite{Dong-TPAMI-2016}. This is the main reason behind our decision to use neural networks. Interestingly, Dong et al.~\cite{Dong-TPAMI-2016} show that deep convolutional neural networks for super-resolution are equivalent to previously-used sparse-coding methods \cite{Yang-TIP-2010}. However, the recent literature indicates that deep neural networks attain better results than handcrafted methods in practice \cite{Dong-TPAMI-2016,Ionescu-CVPR-2019}, mainly because the parameters (weights) are learned from data in an end-to-end fashion. Further specific decisions, such as the neural architecture or the loss function, are taken based on empirical observations.

Our approach is divided into two stages, as illustrated in Figure~\ref{fig_pipeline}. In the first stage, we upsample the image on height and width using a deep fully convolutional neural network. Then, in the second stage, we further upsample the resulting image on the depth axis using another fully convolutional neural network. Therefore, our complete method is designed for resizing the 3D input volume on all three axes. While the CNN used in the first stage resizes the image on two axes, the CNN used in the second stage resizes the image on a single axis. Both CNNs share the same architecture, the only difference being in the upsamling layer (the second CNN upsamples in only one direction). At training time, our CNNs operate on patches. However, since the architecture is fully convolutional, the models can operate on entire slices at inference time, for efficiency reasons.

We hereby note that 3D super-resolution is not equivalent to 2D super-resolution in a slice-by-slice order. Our first CNN performs 2D super-resolution in a slice-by-slice order, increasing an input of size $h \times w \times d$ to the size $r \cdot h \times r \cdot w \times d$, where $r$ is the scale factor. Since we end up with the same number of slices ($r$), this is not enough. This is why we need the second CNN to further increase the image from $r \cdot h \times r \cdot w \times d$ voxels to $r \cdot h \times r \cdot w \times r \cdot d$ voxels. The final output of $r \cdot h \times r \cdot w \times r \cdot d$ voxels could also be obtained by employing a single CNN with 3D convolutions. In most of our convolutional layers, each 2D convolutional filter is formed of $3 \cdot 3 \cdot 32+1=289$ weights to be learned. As we employ two networks for 3D super-resolution, we learn $2 \cdot 289 = 578$ weights. For an equivalent model based on 3D convolutions, each 3D convolutional filter would be formed of $3 \cdot 3 \cdot 3 \cdot 32+1=865$ weights. This analysis proves that our two CNNs put together have less weights than a single 3D CNN. We thus conclude that our approach is more efficient. We note that our approach is essentially based on decomposing the 3D convolutional filters in a product of two 2D convolutional filters. We note that the same principle is applied in literature \cite{Howard-arXiv-2017,Jaderberg-BMVC-2014} to build more efficient CNN models by decomposing 2D convolutional layers in two subsequent 1D convolutional layers that operate on independent dimensions.
 
We further describe in detail the proposed CNN architecture, loss function and data augmentation procedure.

\subsection{Architecture}

Our architecture, depicted in Figure~\ref{fig_architecture} and used for both CNNs, is composed of 10 convolutional (conv) layers, each followed by Rectified Liner Units (ReLU) \cite{Nair-ICML-2010} activations. We decided to use ReLU activations, as this represents the most popular choice of transfer function in current research based on deep learning. All convolutional layers contain filters with a spatial support of $3\times3$. While older deep models were based on larger filters, e.g. the AlexNet \cite{Hinton-NIPS-2012} architecture contains filters of $11\times11$, the recent trend is towards using smaller filters, e.g. the ResNet \cite{He-CVPR-2016} architecture does not contain filters larger than $3\times3$.

Our 10 conv layers are divided into two blocks. The first block, formed of the first 6 conv layers, starts with the input of the neural network and ends just before the upscaling layer. Each of the first 5 convolutional layers are formed of $32$ filters. For the CNN used in the first stage, the number of filters in the sixth convolutional layer is equal to the square of the scale factor, e.g. for a scale factor of $4\times$ the number of filters is $16$. For the CNN used in the second stage, the number of filters in the sixth convolutional layer is equal to the scale factor, e.g. for a scale factor of $4\times$ the number of filters is $4$. The difference is caused by the fact that the first CNN upscales on two axes, while the second CNN upscales on one axis. The first convolutional block contains a short-skip connection, from the first conv layer to the third conv layer, and a long-skip connection, from the first conv layer to the fifth conv layer.

\begin{figure}[t!]
\centering
\includegraphics[width=0.85\linewidth]{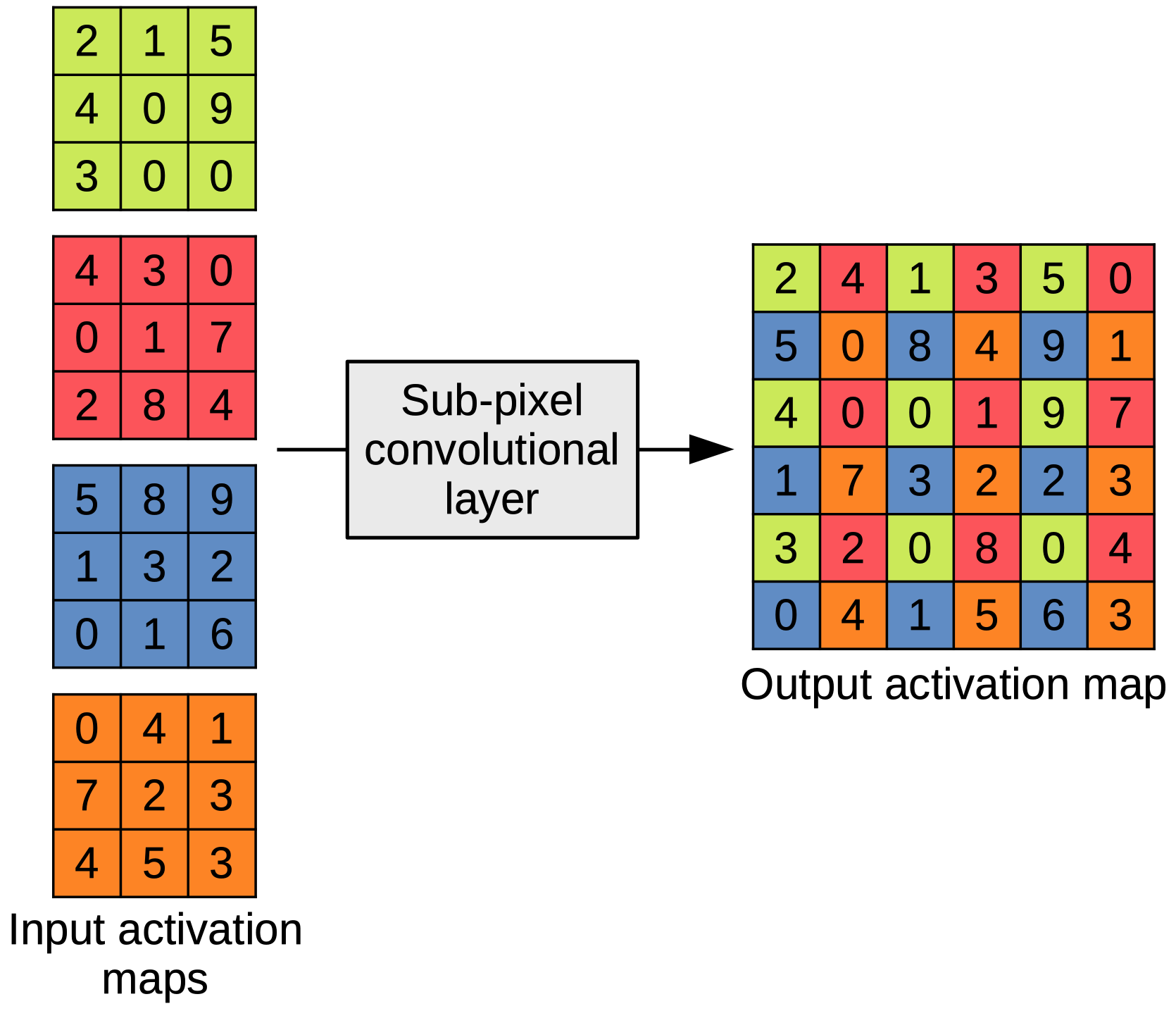}
\caption{An example of low-resolution input activation maps and the corresponding high-resolution output activation map given by the sub-pixel convolutional layer for upscaling on two axes. For a scaling factor of $r=2$ in both directions, the sub-pixel convolutional layer requires $r^2=4$ activation maps as input. Best viewed in color.}
\label{fig_upscaling_two}
\end{figure}

\begin{figure}[t!]
\centering
\includegraphics[width=0.85\linewidth]{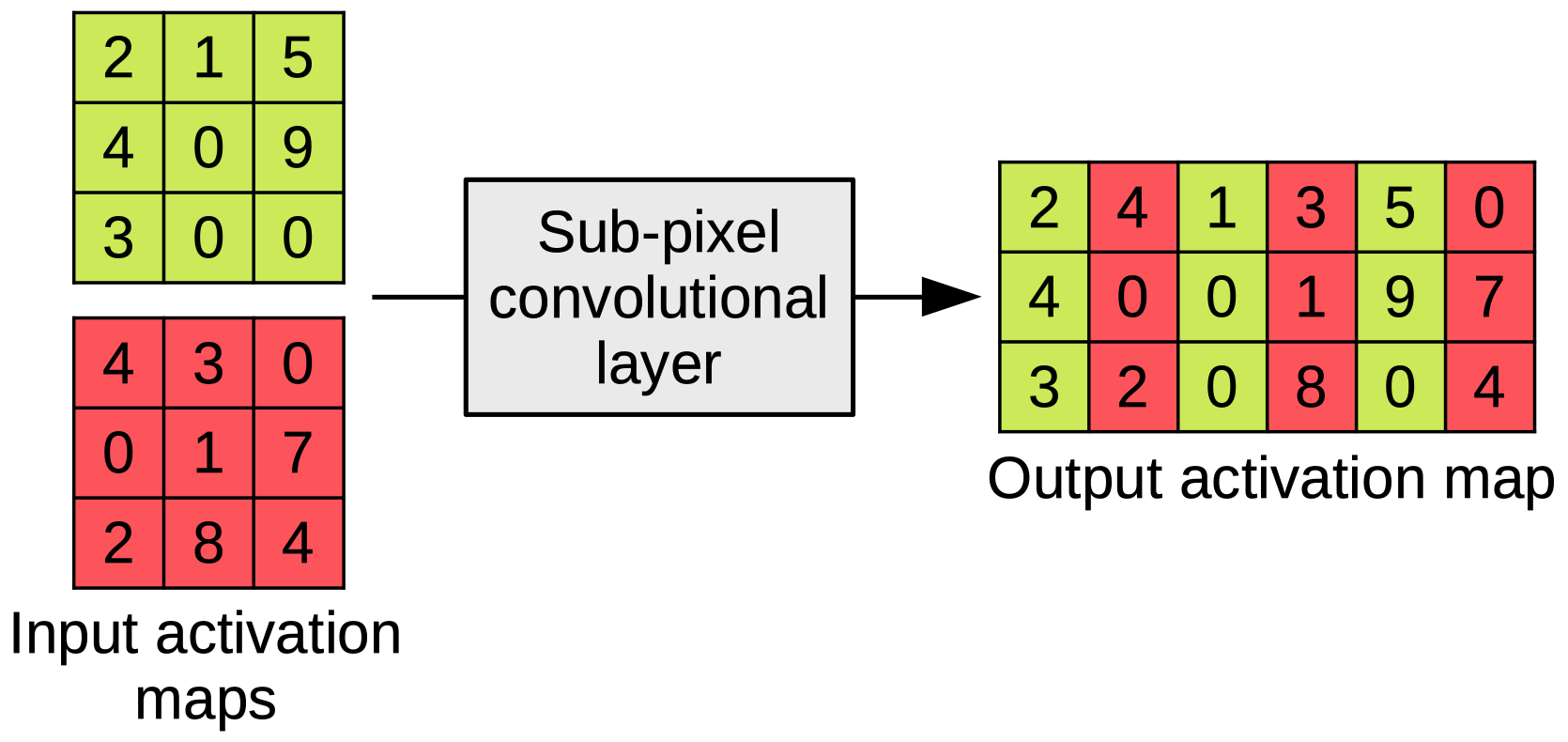}
\caption{An example of low-resolution input activation maps and the corresponding high-resolution output activation map given by the sub-pixel convolutional layer for upscaling on one axis. For a scaling factor of $r=2$ in one direction, the sub-pixel convolutional layer requires $r=2$ activation maps as input. Best viewed in color.}
\label{fig_upscaling_one}
\end{figure}

The first convolutional block is followed by a sub-pixel convolutional (upscaling) layer, which was introduced in \cite{Shi-CVPR-2016}. In the upscaling layer, the activation maps produced by the sixth conv layer are assembled into a single activation map. Throughout the first convolutional block, the spatial size of the low-resolution input is preserved, i.e. the activation maps of the sixth conv layer have $h_I \times w_I$ components, where $h_I$ and $w_I$ are the height and the width of the input image $I$. In order to increase the input $r$ times on both axes, the output of the sixth conv layer must be a tensor of $h_I \times w_I \times r^2$ components. The activation map resulting from the sub-pixel convolutional layer is a matrix of $(h_I \cdot r) \times (w_I \cdot r)$ components. For super-resolution on two axes, the pixels are rearranged as shown in Figure~\ref{fig_upscaling_two}. In a similar fashion, we can increase the input $r$ times on one axis. In this case, the output of the sixth conv layer must be a tensor of $h_I \times w_I \times r$ components. This time, the activation map resulting from the sub-pixel convolutional layer can be either a matrix of $(h_I \cdot r) \times w_I$ components or a matrix of $h_I \times (w_I \cdot r)$ components, depending on the direction we aim to increase the resolution. For super-resolution on one axis, the pixels are rearranged as shown in Figure~\ref{fig_upscaling_one}. To our knowledge, we are the first to propose a sub-pixel convolutional (upscaling) layer for super-resolution in one direction.

When Shi et al.~\cite{Shi-CVPR-2016} introduced the sub-pixel convolutional layer, they used it as the last layer of their CNN. Hence, the output depth of the upscaling layer is equal to the number of channels in the output image. Different from Shi et al.~\cite{Shi-CVPR-2016}, we employ further convolutions after the upscaling layer. Nevertheless, since we are working with CT/MRI (grayscale) images, our final output has a single channel. In our architecture, the upscaling layer is followed by our second convolutional block, which starts with the seventh convolutional layer and ends with the tenth convolutional layer. The first three conv layers in our second block are formed of $32$ filters. The tenth conv layer contains a single convolutional filter, since our output is a grayscale image that has a single channel. The second convolutional block contains a short skip connection, from the seventh conv layer to the ninth conv layer. The spatial size of $h_O \times w_O$ components of the activation maps is preserved throughout the second convolutional block, where $h_O$ and $w_O$ are the height and the width of the output image $O$.

\subsection{Losses and Optimization}

In order to obtain a CNN model for single-image super-resolution, the aim is to minimize the differences between the ground-truth high-resolution image and the output image provided by the CNN. Researchers typically employ the mean absolute difference as the objective function. Given a low-resolution input image $I$ and the corresponding ground-truth high-resolution image $O$, the loss based on the mean absolute value is formally defined as follows:
\begin{equation}\label{eq_loss_std}
\mathcal{L}(\theta,I,O) = \sum_{i=1}^{w_O} \sum_{j=1}^{h_O} | f(\theta,I) - O |,  
\end{equation}
where $\theta$ are the CNN parameters (weights), $f$ is the transformation function learned by the CNN, and $w_O$ and $h_O$ represent the width and the height of the ground-truth image $O$, respectively. 

When we train our CNN model, we do not employ the standard approach of minimizing the difference between the output provided by the CNN and the ground-truth HR image. Instead, we propose a novel approach based on an intermediate loss function. Since the conv layers after the upscaling layer are meant to refine the high-resolution image without taking any additional information from the low-resolution input image, we note that the high-resolution image resulting immediately after the upscaling layer should be as similar as possible to the ground-truth high-resolution image. Therefore, we propose a loss function that aims to minimize the difference between the intermediately-obtained HR image and the ground-truth HR image, in addition to minimizing the difference between the final HR output image and the ground-truth HR image. Let $f_1$ denote the transformation function that corresponds to the first convolutional block and the upscaling layer, and let $f_2$ denote the transformation function that corresponds to the second convolutional block. With these notations, the transformation function $f$ of our full CNN architecture can be written as follows:
\begin{equation}
f(\theta,I) = f_2(\theta_2, f_1(\theta_1, I)),
\end{equation}
where $\theta$ are the parameters of the full CNN, $\theta_1$ are the parameters of the first convolutional block and $\theta_2$ are the parameters of the second convolutional block, i.e. $\theta$ is a concatenation of $\theta_1$ and $\theta_2$. Having defined $f_1$ and $f_2$ as above, we can formally write our loss function as follows:
\begin{equation}\label{eq_loss_full}
\mathcal{L}_{full} = \mathcal{L}_{standard} + \lambda \cdot \mathcal{L}_{intermediate},
\end{equation}
where $\lambda$ is a parameter that controls the importance of the intermediate loss with respect to the standard loss, $\mathcal{L}_{standard}$ is the standard loss defined in Equation~\eqref{eq_loss_std} and $\mathcal{L}_{intermediate}$ is defined as follows:
\begin{equation}\label{eq_loss_intermediate}
\mathcal{L}_{intermediate}(\theta_1,I,O) = \sum_{i=1}^{w_O} \sum_{j=1}^{h_O} | f_1(\theta_1,I) - O |.
\end{equation}

In the experiments, we set $\lambda=1$, since we did not find any strong reason to assign a lower or higher importance to the intermediate loss. By replacing $\lambda$ with $1$ and the loss values from Equation~\eqref{eq_loss_std} and Equation~\eqref{eq_loss_intermediate}, Equation~\eqref{eq_loss_full} becomes:
\begin{equation}\label{eq_loss_full_explicit}
\begin{split}
\mathcal{L}_{full}(\theta,I,O) &= \sum_{i=1}^{w_O} \sum_{j=1}^{h_O} | f(\theta,I) - O | \\
&+  \sum_{i=1}^{w_O} \sum_{j=1}^{h_O} | f_1(\theta_1,I) - O |.
\end{split}
\end{equation}

In order to optimize towards the objective defined in Equation~\eqref{eq_loss_full_explicit}, we employ the Adam optimizer \cite{Kingma-ICLR-2015}, which is known to converge faster than Stochastic Gradient Descent.

\subsection{Data Augmentation}

\begin{figure}[t!]
\centering
\includegraphics[width=0.8\linewidth]{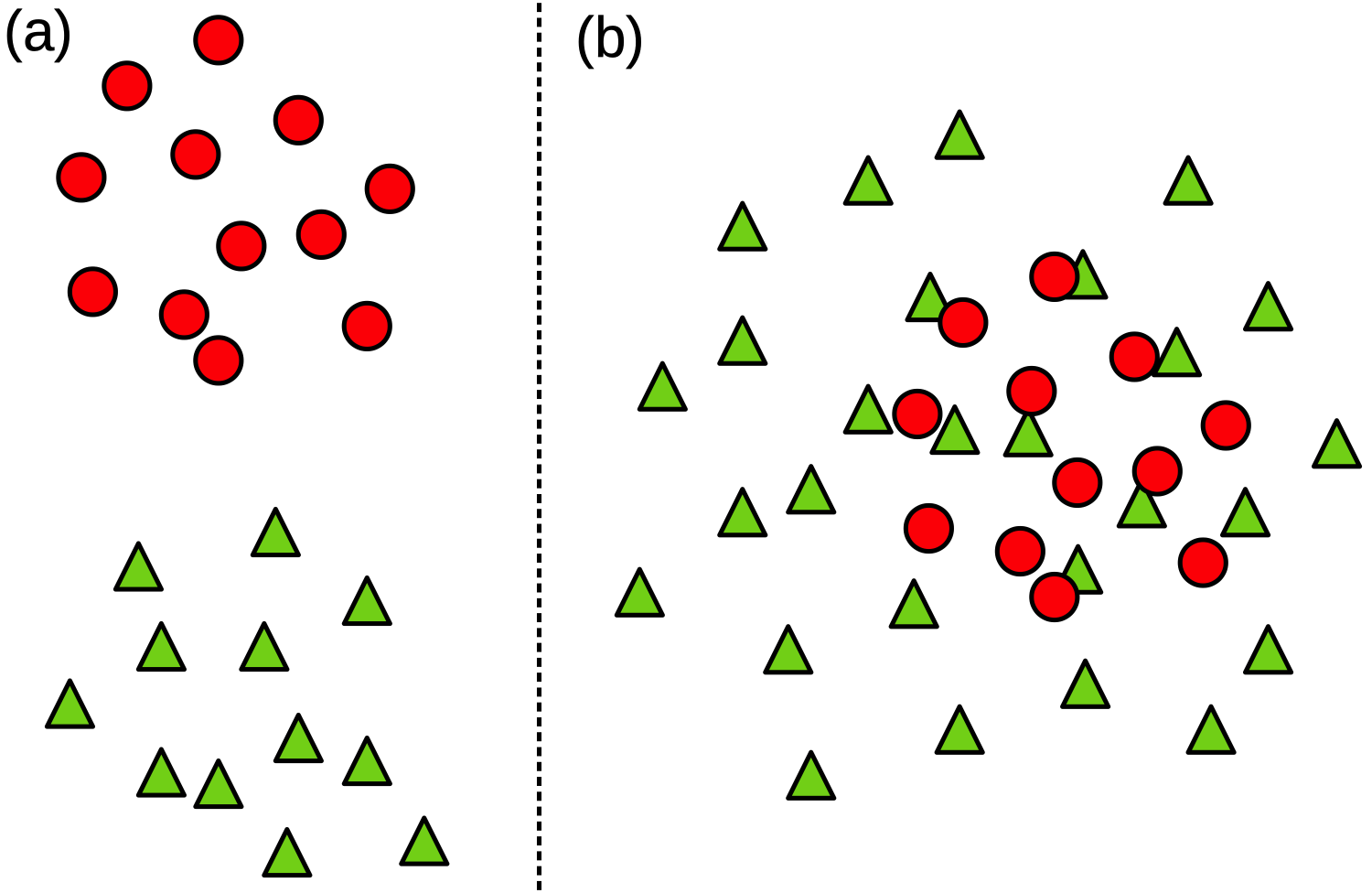}
\caption{Distribution of training samples (represented by green triangles) and test samples (represented by red circles), when the training samples are smoothed using a fixed standard deviation (left-hand side) versus using a randomly-chosen standard deviation (right-hand side). In example $(a)$, overfitting on the training data leads to poor results on test data. In example $(b)$, the danger of overfitting is diminished because the test distribution is included in the training distribution. Best viewed in color.}
\label{fig_data_blur}
\end{figure}

A common approach to force the CNN to produce sharper output images is to apply Gaussian smoothing using a fixed standard deviation at training time \cite{Du-AS-2019, Shi-JBHI-2018, ZENG-CBM-2018}. By training the CNN on blurred low-resolution images, the super-resolution task becomes harder. During inference, when the input images are no longer blurred, the task will be much easier. However, smoothing only the training images with a fixed standard deviation will inherently generate a distribution gap between training and test data. If the CNN fits the training distribution well, it might not produce the desired results at inference time. This is because machine learning models are based on the assumption that training data and test data are sampled from the same distribution. We propose to solve this problem by using a randomly-chosen standard deviation for each training image. Although the training data distribution will still be different from the testing data distribution, it will include the distribution of test samples, as illustrated in Figure~\ref{fig_data_blur}.  In order to augment the training data, we apply Gaussian blur with a probability of $0.5$ (only half of the images are smoothed) using a kernel of $3 \times 3$ components and a randomly-chosen standard deviation between $0$ and $0.5$. In this way, we increase the variance of the training data without introducing any bias. In this case, if the CNN fits the training distribution well, it will produce good super-resolution outputs during inference, since there is no distribution gap between training and test.

\section{Experiments}
\label{sec_Experiments}

\subsection{Data Sets}

The first data set used in the experiments consists of $10$ anonymized 3D images of brain CT provided by the Medical School at Colțea Hospital. We further refer to this data set as the Colțea Hospital (CH) data set. In order to fairly train and test our CNN models and baselines, we randomly selected $6$ images for training and used the remaining $4$ images for testing. The training set has $359$ slices (2D images) and the testing set has $238$ slices. The height and the width of the slices vary between $192$ and $512$ pixels, while the depth of the 3D images varies between $3$ and $176$ slices. The resolution of a voxel is $1 \times 1 \times 1\; mm^3$.

The second data set used in our experiments is the National Alliance for Medical Image Computing (NAMIC) Brain Multimodality data set. The NAMIC data set consists of $20$ 3D MRI images, each composed of $176$ slices of $256 \times 256$ pixels. As for the CH data set, the resolution of a voxel is $1 \times 1 \times 1\;mm^3$. For our experiments, we used T1-weighted (T1w) and T2-weighted (T2w) images independently. Following \cite{Pham-CMIG-2019}, we split the NAMIC data set into a training set containing $10$ 3D images and a test set containing the other $10$ images. We kept the same split for T1w and T2w.

\subsection{Experimental Setup}

\subsubsection{Evaluation Metrics}

As most previous works \cite{ Du-AS-2019,  Du-NC-2019,  Li-Access-2019, Oktay-MICCAI-2016, Pham-CMIG-2019, Shi-JBHI-2018, You-TMI-2019, Yu-ICIP-2017, ZENG-CBM-2018, Zhao-TMI-2019}, we employed the two most common evaluation metrics, namely the peak signal-to-noise ratio (PSNR) and the structural similarity index (SSIM). The PSNR is the ratio between the maximum possible power of a signal and the power of corrupting noise that affects the fidelity of its representation. Although the PSNR is one of the most used metrics for image reconstruction, some researchers \cite{Wang-TIP-2004,Wang-SPM-2009} argued that it is not highly indicative of the perceived similarity. The SSIM \cite{Wang-TIP-2004} aims to address this shortcoming by taking contrast, luminance and texture into account. The result of the SSIM is a number between -1 and 1, where a value of 1 means the ground-truth image and the reconstructed image are identical. Similarly, a higher PSNR value indicates a better reconstruction, although the PSNR does not have an upper bound.

Since PSNR and SSIM values cannot guarantee a visually favorable result, we employ an additional metric for the final results, namely the information fidelity criterion (IFC) \cite{Sheikh-TIP-2005}. Although IFC is scarcely used in literature \cite{You-TMI-2019}, Yang et al.~\cite{Yang-ECCV-2014} pointed out that IFC is correlated well with the human perception of SR images. As for PSNR and SSIM, higher IFC values indicate better results.

\subsubsection{Image Quality Assessment by Human Observers}

Because the above metrics rely only on the pixel values and not on the perceived visual quality, we decided to evaluate our method with the help of human annotators. Although a deep learning method can provide better PSNR, SSIM or IFC values, it might produce artifacts that could be misleading for right diagnostics and treatment. We thus have to make sure that our approach does not produce any unwanted artifacts visible to humans. We conducted the image quality assessment on the CH data set, testing our CNN-based method against Lanczos interpolation. We used CT slices extracted from high-resolution 3D images resulting after applying super-resolution on all three axes. For each upsampling factor, $2\times$ and $4\times$, we extracted 100 CT slices at random from the test set. Hence, each human evaluator had to annotate 200 image pairs (100 for each upsampling factor). For each evaluation sample, an annotator would have seen the original image in the middle and the two reconstructed images on its sides, one on the left side and the other on the right side. The annotators had a magnifying glass tool that allowed them to look at details and discover artifacts. The locations (left or right) of the images reconstructed by our CNN and by Lanczos interpolation were randomly picked every time. To prevent any form of cheating, the randomly picked locations were unknown to the annotators. For each test sample, we asked each annotator to select the image that best reconstructed the original image. Our experiment was completed by 18 human annotators, 6 of them being doctors specialized in radiotherapy and oncology. In total, we collected $3600$ annotations (18 annotators $\times$ 200 samples).

\subsubsection{Baselines}

We compared our method with standard resizing methods based on various interpolation schemes, namely nearest neighbors, bilinear, bicubic and Lanczos. In addition to these baselines, we compared with three methods \cite{Du-AS-2019, You-TMI-2019, ZENG-CBM-2018} that focused on 2D SISR and one method \cite{Pham-CMIG-2019} that focused on 3D SISR. We note that You et al.~\cite{You-TMI-2019} did not report results on NAMIC. Nonetheless, You et al.~\cite{You-TMI-2019} were kind to provide access to their source code. We thus applied their method on both CH and NAMIC data sets, keeping the same settings and hyperparameters as recommended by the authors. As You et al.~\cite{You-TMI-2019}, we employed their method only on 2D super-resolution for the $2\times$ upscaling factor. For the other three baselines, we included the NAMIC scores reported in the respective articles \cite{Du-AS-2019, Pham-CMIG-2019, ZENG-CBM-2018}.

\begin{table}[!t]
\caption{Preliminary 2D super-resolution results on the CH data set for an upscaling factor of $2\times$. The PSNR and the SSIM values are reported for various patch sizes and different numbers of filters. For models with $7 \times 7$ patches, we report the inference time (in seconds) per CT slice measured on an Nvidia GeForce 940MX GPU with 2GB of RAM.}\label{tab_preliminary_results}
\begin{center}
\setlength\tabcolsep{3.0pt}
\begin{tabular}{|c|c|c|c|c|}
\hline  
Number of filters& Input size     & SSIM           & PSNR       & Time (in seconds)      \\ 
\hline
\hline   
$32$             &  $4 \times 4$  & $0.9165$       & $35.36$    &         -              \\  
$32$             &  $7 \times 7$  & $0.9270$       & $36.22$    &        0.05             \\   
$32$             & $14 \times 14$ & $0.8987$       & $33.83$    &         -              \\  
$32$             & $17 \times 17$ & $0.8934$       & $33.42$    &         -              \\  
\hline   
$64$             &  $7 \times 7$  & $0.9279$       & $36.28$    &        0.08            \\ 
$128$            &  $7 \times 7$  & $0.9276$       & $36.28$    &        0.16            \\  
\hline   
\end{tabular}
\end{center}
\end{table}

\subsection{Parameter Tuning and Preliminary Results} 

We conducted a series of preliminary experiments to determine the optimal patch size as well as the optimal width (number of convolutional filters) for our CNN. In order to find the optimal patch size, we tried out patches of $4\times4$, $7\times7$, $14\times14$ and $17\times17$ pixels. In term of the number of filters, we tried out values in the set $\{32, 64, 128\}$ for all conv layers in our network. These parameters were tuned in the context of 2D super-resolution on the CH data set. The corresponding results are presented in Table~\ref{tab_preliminary_results}. First of all, we note that our method produces better SSIM and PSNR values, i.e. $0.9270$ and $36.22$, for patches of $7 \times 7$ pixels. Second of all, we observe that adding more filters on the conv layers slightly increases the SSIM and PSNR values. However, the gains in terms of SSIM and PSNR come with a great cost in terms of time. For example, using $128$ filters on each conv layer triples the processing time in comparison with using $32$ filters on each conv layer. For the subsequent experiments, we thus opted for patches of $7 \times 7$ pixels and conv layers with $32$ filters.

We believe that it is important to note that, although the number of training CT slices is typically in the range of a few hundreds, the number of training patches is typically in the range of hundreds of thousands. For instance, the number of $7 \times 7$ training patches extracted from the CH data set for the $2\times$ upscaling factor is $326,\!000$. We thus stress out that the number of training samples is high enough to train highly-accurate deep learning models.

During training, we used mini-batches of $128$ images throughout all the experiments. In a set of preliminary experiments, we did not observe any significant differences when using mini-batches of $64$ or $256$ images. In each experiment, we trained the CNN for $40$ epochs, starting with a learning rate (step size) of $10^{-3}$ and decreasing the learning rate to $10^{-4}$ after the first $20$ epochs.

\subsection{Ablation Study Results}

\begin{table*}[!t]
\caption{Ablation 2D super-resolution results on the CH data set for an upscaling factor of $2\times$. The PSNR and the SSIM values are reported for various ablated versions of our CNN model. The best results are highlighted in bold. Results of ablated models marked with $\dagger$ are significantly worse than our complete model, according to paired McNemar's testing \cite{Dietterich-NC-1998} for the significance level $0.001$.}\label{tab_ablation_results}
\begin{center}
\begin{tabular}{|c|c|c|c|c|c|c|}
\hline  
Second conv block & Intermediate loss  & Short-skip connections & Long-skip connection & Variable standard deviation & SSIM           & PSNR   \\  
\hline
\hline
    \xmark     &   \xmark          &     \xmark            &      \xmark          &      \xmark                   & $0.9224^{\dagger}$       & $35.58^{\dagger}$\\

\hline   
    \xmark     &   \xmark          &     \cmark            &      \cmark          &      \cmark                   & $0.9236^{\dagger}$       & $35.94^{\dagger}$\\  
\hline
    \cmark     &   \xmark          &     \cmark            &      \cmark          &      \cmark                   & $0.9256^{\dagger}$       & $36.15^{\dagger}$\\   
\hline   
    \cmark     &   \cmark          &     \xmark            &      \cmark          &      \cmark                   & $0.9260^{\dagger}$       & $36.17^{\dagger}$\\
\hline   
    \cmark     &   \cmark          &     \cmark            &      \xmark          &      \cmark                   & $0.9234^{\dagger}$       & $36.11^{\dagger}$\\ 
\hline 
    \cmark     &   \cmark          &     \cmark            &      \cmark          &      \xmark                   & $0.9236^{\dagger}$       & $35.69^{\dagger}$\\    
\hline
    \cmark     &   \cmark          &     \cmark            &      \cmark          &      \cmark                   & $\mathbf{0.9270}$       	& $\mathbf{36.22}$\\  
  
\hline   
\end{tabular}
\end{center}
\end{table*}

We performed an ablation study to emphasize the effect of various components over the overall performance. The ablation results obtained on the CH data set for super-resolution on height and width by a factor of $2\times$ are presented in Table~\ref{tab_ablation_results}.

In our first ablation experiment, we have eliminated all the enhancements in order to show the performance level of a baseline CNN on the CH data set. Since there are several SISR works \cite{Du-AS-2019, Hatvani-TRPMS-2018, Yu-ICIP-2017, Zhao-TMI-2019} based on the standard ESPCN model \cite{Shi-CVPR-2016}, we have eliminated the second convolutional block in the second ablation experiment, transforming our architecture into a standard ESPCN architecture. The performance drops from $0.9270$ to $0.9236$ in terms of SSIM and from $36.22$ to $35.94$ in terms of PSNR. In the subsequent ablation experiments, we have removed, in turns, the intermediate loss, the short-skip connections and the long-skip connection. The results presented in Table~\ref{tab_ablation_results} indicate that all these components are relevant to our model, bringing significant performance benefits in terms of both SSIM and PSNR. In our last ablation experiment, we used a fixed standard deviation instead of a variable one for the Gaussian blur added on training patches. We notice that our data augmentation approach based on a variable standard deviation brings the highest gains in terms of SSIM (from $0.9236$ to $0.9270$) and PSNR (from $35.69$ to $36.22$), with respect to the other ablated components. 

Since the differences in terms of PSNR or SSIM for the ablated models are hard to quantify as small or large with respect to the complete CNN, we conducted paired McNemar's significance testing \cite{Dietterich-NC-1998} to determine if the differences are statistically significant or not. We considered a p-value threshold of $0.001$ for our statistical testing. Every ablated model that is significantly different from the complete model is marked with $\dagger$ in Table~\ref{tab_ablation_results}. We note that the complete CNN is significantly better than each ablated version, although the actual differences in terms of PSNR or SSIM might seem small. We thus conclude that all the proposed enhancements provide significant performance gains.

\subsection{Results on CH Data Set}

\begin{table*}[!t]
\caption{2D super-resolution results of our CNN model versus a state-of-the-art method \cite{You-TMI-2019} and several interpolation baselines on the CH data set. The PSNR, the SSIM and the IFC values are reported for two upscaling factors, $2\times$ and $4\times$. The best result on each column is highlighted in bold.}\label{tab_ch_results_2d}
\begin{center}
\begin{tabular}{|l|c|c|c|c|c|c|}
\hline 
Method						& \multicolumn{3}{|c|}{$2\times$}					& \multicolumn{3}{|c|}{$4\times$} \\
\cline{2-7}				& SSIM				& PSNR			& IFC           			& SSIM				& PSNR			& IFC\\
\hline
\hline
Nearest neighbor		& $0.8893$ 		& $32.71$	& $4.40$				& $0.7659$		& $29.06$	& $1.32$\\ 
\hline  
Bilinear						& $0.8835$		& $33.34$	& $3.73$				& $0.7725$		& $29.73$	& $1.49$\\ 
\hline  
Bicubic						& $0.9077$		& $34.71$	& $4.59$				& $0.7965$		& $30.41$	& $1.72$\\ 
\hline   
Lanczos					& $0.9111$		& $35.08$	& $4.93$				& $0.8012$		& $30.57$	& $1.84$\\  
\hline
You et al.~\cite{You-TMI-2019} & $0.8874$		& $32.73$	& $4.40$		& -					& -				& -\\  
\hline 
Our CNN model		& $\mathbf{0.9291}$ & $\mathbf{36.39}$	& $\mathbf{5.36}$ & $\mathbf{0.8308}$ & $\mathbf{31.59}$ & $\mathbf{1.92}$\\ 
\hline 
\end{tabular}
\end{center} 
\end{table*}

\begin{table*}[!t]
\caption{3D super-resolution results of our CNN model versus several interpolation baselines on the CH data set. The PSNR, the SSIM and the IFC values are reported for two upscaling factors, $2\times$ and $4\times$. The best result on each column is highlighted in bold.}\label{tab_ch_results_3d}
\begin{center}
\begin{tabular}{|l|c|c|c|c|c|c|}
\hline 
Method & \multicolumn{3}{|c|}{$2\times$} & \multicolumn{3}{|c|}{$4\times$} \\
\cline{2-7}				& SSIM				& PSNR			& IFC           			& SSIM				& PSNR			& IFC\\
\hline  
\hline  
Nearest neighbor		& $0.8430$		& $30.36$	& $2.19$				& $0.7152$		& $27.32$	& $0.72$\\ 
\hline  
Bilinear						& $0.8329$		& $30.72$	&	$2.18$				& $0.7206$		& $27.93$ 	& $0.95$\\ 
\hline  
Bicubic						& $0.8335$		& $26.51$	&	$2.47$				& $0.7200$		& $24.05$ 	& $1.04$\\ 
\hline   
Lanczos					& $0.8423$		& $27.85$	&	$2.58$				& $0.7263$		& $25.06$	& $1.09$\\  
\hline 
Our CNN model		& $\mathbf{0.8926}$ & $\mathbf{33.04}$ & $\mathbf{2.83}$	& $\mathbf{0.7819}$ & $\mathbf{29.36}$ & $\mathbf{1.13}$\\ 
\hline 
\end{tabular}
\end{center}
\end{table*}

We first compared our CNN-based model with a series of interpolation baselines and a state-of-the-art method \cite{You-TMI-2019} on the CH data set. We present the results for super-resolution on two axes (width and height) in Table~\ref{tab_ch_results_2d}. Among the considered baselines, it seems that the Lanczos interpolation method provides better results than the bicubic, the bilinear or the nearest neighbor methods. Our CNN model is able to surpass all baselines for both upscaling factors, $2\times$ and $4\times$. Compared to the best interpolation method (Lanczos), our method is $0.0180$ better in terms of SSIM, $1.31$ better in terms of PSNR and $0.43$ better in terms of IFC. Furthermore, our CNN provides superior results to the GAN-based method of You et al.~\cite{You-TMI-2019}.

We note that, in Table~\ref{tab_ablation_results}, we reported an SSIM of $0.9270$ and a PSNR of $36.22$ for our method, while in Table~\ref{tab_ch_results_2d}, we reported an SSIM of $0.9291$ and a PSNR of $36.39$. In order to boost the performance of our method in accordance with the observed differences between Tables~\ref{tab_ablation_results} and~\ref{tab_ch_results_2d}, we employed the self-ensemble strategy used by Lim et al.~\cite{Lim-CVPRW-2017}. For each input image, the self-ensemble strategy consists in generating additional images using geometric transformations, e.g. rotations and flips. Following Lim et al.~\cite{Lim-CVPRW-2017}, we generated 7 augmented images from the LR input image, upsampling all 8 images (the original image and the 7 additional ones) using our CNN. We then applied the inverse transformations to the resulting 8 HR images in order to obtain 8 output images that are aligned with the ground-truth HR images. The final output image is obtained by taking the median of the HR images. In the following experiments on CH and NAMIC data sets, the reported results always include the described self-ensemble strategy.

\begin{table*}[!t]
\caption{2D super-resolution results of our CNN model versus several state-of-the-art methods \cite{Du-AS-2019,You-TMI-2019,ZENG-CBM-2018} and the Lanczos interpolation baseline on the NAMIC data set. For Zeng et al.~\cite{ZENG-CBM-2018}, we included results for both single-channel super-resolution (SCSR) and multi-channel super-resolution (MCSR). The PSNR, the SSIM and the IFC values are reported for both T1w and T2w images and for two upscaling factors, $2\times$ and $4\times$. The best results on each column are highlighted in bold.}\label{tab_namic_results_2D}
\begin{center}
\begin{tabular}{|l|c|c|c|c|c|c|c|c|c|c|c|c|}
\hline 
  & \multicolumn{6}{|c|}{T1-weighted} & \multicolumn{6}{|c|}{T2-weighted}\\
\cline{2-13}
Method        							& \multicolumn{3}{|c|}{$2\times$} & \multicolumn{3}{|c|}{$4\times$} & \multicolumn{3}{|c|}{$2\times$} & \multicolumn{3}{|c|}{$4\times$}\\
\cline{2-13}   												& SSIM         & PSNR      & IFC			& SSIM        & PSNR & IFC    & SSIM           & PSNR & IFC          & SSIM           & PSNR & IFC\\
\hline  
\hline   
Lanczos														& $0.9635$ & $37.03$ & $3.06$ & $0.8955$ & $32.71$  & $1.28$ & $0.9788$ & $39.64$ & $3.22$ & $0.9143$ & $33.80$ & $1.37$\\
\hline
Zeng et al.~\cite{ZENG-CBM-2018} (SCSR)	& $0.9220$ & $36.86$ & $-$ & $0.7120$ & $28.33$ & $-$ & $-$  & $-$ & $-$  & $-$ & $-$  & $-$\\  
\hline 
Zeng et al.~\cite{ZENG-CBM-2018} (MCSR)	& $0.9450$ & $38.32$ & $-$ & $0.8110$ & $30.84$ & $-$ & $-$  & $-$ & $-$  & $-$ & $-$  & $-$\\  
\hline 
Du et al.~\cite{Du-AS-2019}						& $0.9390$ & $37.21$ & $-$ & $0.7370$ & $29.05$ & $-$ & $-$  & $-$ & $-$  & $-$ & $-$  & $-$\\
\hline
You et al.~\cite{You-TMI-2019}  				& $0.9448$  & $35.09$ & $2.80$  		& $-$ & $-$  & $-$    	& $0.9594$  & $36.19$ & $2.89$  & $-$  & $-$ & $-$\\
\hline 
Our CNN model            & $\mathbf{0.9775}$ & $\mathbf{39.29}$ & $\mathbf{3.56}$	& $\mathbf{0.9193}$ & $\mathbf{33.74}$ & $\mathbf{1.29}$ & $\mathbf{0.9882}$ & $\mathbf{42.20}$ & $\mathbf{3.79}$ & $\mathbf{0.9382}$ & $\mathbf{34.86}$ & $\mathbf{1.39}$\\ 
\hline 
\end{tabular}
\end{center}
\end{table*}

\begin{table*}[!t]
\caption{3D super-resolution results of our CNN model versus a state-of-the-art method \cite{Pham-CMIG-2019} and the Lanczos interpolation baseline on the NAMIC data set. The PSNR, the SSIM and the IFC values are reported for both T1w and T2w images and for two upscaling factors, $2\times$ and $4\times$. The best results on each column are highlighted in bold.}\label{tab_namic_results_3D}
\begin{center}
\begin{tabular}{|l|c|c|c|c|c|c|c|c|c|c|c|c|}
\hline 
  & \multicolumn{6}{|c|}{T1-weighted} & \multicolumn{6}{|c|}{T2-weighted} \\
\cline{2-13}
Method        & \multicolumn{3}{|c|}{$2\times$} & \multicolumn{3}{|c|}{$4\times$} & \multicolumn{3}{|c|}{$2\times$} & \multicolumn{3}{|c|}{$4\times$} \\
\cline{2-13}   												& SSIM         & PSNR      & IFC			& SSIM        & PSNR & IFC    & SSIM           & PSNR & IFC          & SSIM           & PSNR & IFC\\
\hline  
\hline   
Lanczos               								& $0.9423$ & $35.72$ & $2.00$ & $0.8690$ & $31.81$ & $0.95$ & $0.9615$ & $37.80$ & $2.29$ & $0.8829$ & $32.08$ & $1.03$\\
\hline
Pham et al.~\cite{Pham-CMIG-2019}  & $-$  & $-$ & $-$  & $-$ & $-$  & $-$   & $0.9781$ & $38.28$ & $-$  & $-$ & $-$ & $-$\\
\hline 
Our CNN model            						& $\mathbf{0.9687}$ & $\mathbf{37.85}$ & $\mathbf{2.38}$ & $\mathbf{0.9050}$ & $\mathbf{32.88}$ & $\mathbf{0.99}$ & $\mathbf{0.9835}$ & $\mathbf{40.57}$ & $\mathbf{2.67}$ & $\mathbf{0.9251}$ & $\mathbf{33.54}$ & $\mathbf{1.10}$\\ 
\hline 
\end{tabular}
\end{center}
\end{table*}

\begin{figure*}[t!]
\centering
\includegraphics[width=0.95\linewidth]{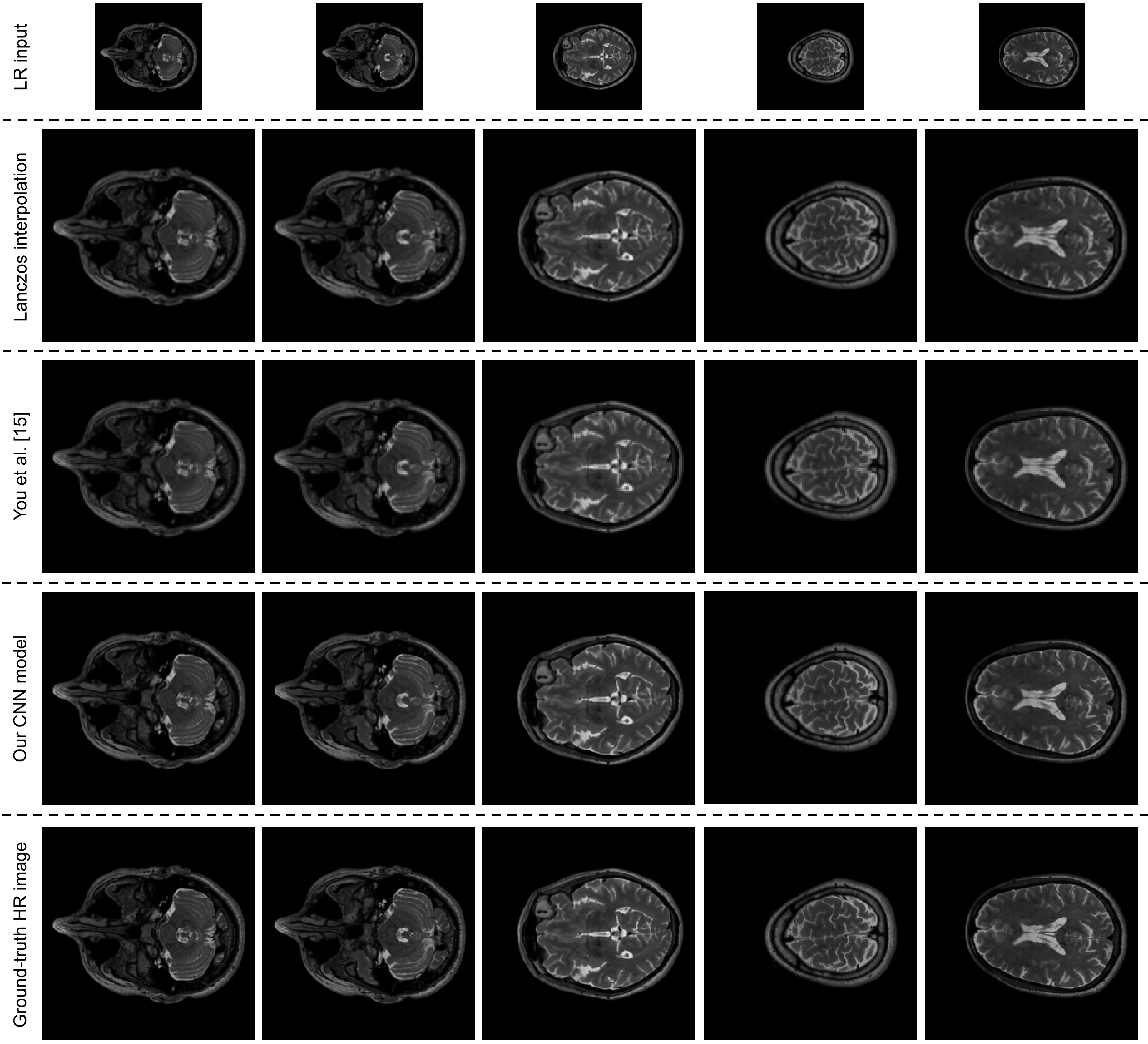}
\caption{Image super-resolution examples selected from the NAMIC data set. In order to obtain the input images of $128 \times 128$ pixels, the original NAMIC images were downsampled by a scale factor of $2\times$. HR images of $256 \times 256$ pixels generated by Lanczos interpolation, by the GAN-based method of You et al.~\cite{You-TMI-2019} and by our CNN model are compared with the original (ground-truth) HR images.}
\label{fig_namic_results}
\end{figure*}

We provide the results for super-resolution on all three axes in Table~\ref{tab_ch_results_3d}. First of all, we notice that the SSIM, the PSNR and the IFC values are lower for all methods when dealing with 3D super-resolution (Table~\ref{tab_ch_results_3d}) instead of 2D super-resolution (Table~\ref{tab_ch_results_2d}). This shows that the task of 3D super-resolution is much harder than 2D super-resolution. This is an expected result, considering that the dimensionality of the reconstruction space increases significantly for 3D super-resolution, i.e. there are many more HR outputs corresponding to a single LR input, while the training data is the same. Nevertheless, our method exhibits smaller performance drops when going from 2D super-resolution to 3D super-resolution. As for the 2D super-resolution experiments on CH data set, our CNN model for 3D super-resolution is superior to all baselines for both upscaling factors. We thus conclude that our CNN model is better than all interpolation baselines on the CH data set, for both 2D and 3D super-resolution and for all upscaling factors.

\subsection{Results on NAMIC Data Set}

On the NAMIC data set, we compared our method with the best-performing interpolation method on the CH data set, namely Lanczos interpolation, as well as some state-of-the-art 2D \cite{Du-AS-2019,You-TMI-2019,ZENG-CBM-2018} and 3D \cite{Pham-CMIG-2019} super-resolution methods. We note that most previous works, including \cite{Du-AS-2019,Pham-CMIG-2019,ZENG-CBM-2018}, used bicubic interpolation as a relevant baseline. Unlike these works, we opted for Lanczos interpolation, which provided better results than bicubic interpolation and other interpolation methods on the CH data set.

We first present the 2D super-resolution results in Table~\ref{tab_namic_results_2D}. The 2D SR results indicate that the GAN-based method of You et al.~\cite{You-TMI-2019} is superior to the CNN baselines \cite{Du-AS-2019,ZENG-CBM-2018}. However, none of the state-of-the-art methods \cite{Du-AS-2019,You-TMI-2019,ZENG-CBM-2018} is able to attain better performance than Lanczos interpolation. This proves that Lanczos interpolation is a much stronger baseline. Among the deep learning methods, our CNN is the only one to surpass Lanczos interpolation for 2D SR on NAMIC. We believe that this result is noteworthy.

We also present the 3D super-resolution results in Table~\ref{tab_namic_results_3D}. The 3D SR results show that the approach of Pham et al.~\cite{Pham-CMIG-2019} is better than Lanczos interpolation, which is remarkable. Our CNN is even better, surpassing both the Lanczos interpolation and the approach of Pham et al.~\cite{Pham-CMIG-2019}.

As for the CH data set, we observe that the PSNR, the SSIM and the IFC scores for 2D super-resolution are higher than the corresponding scores for 3D super-resolution. The same explanation applies to the NAMIC data set, i.e. the CNNs have to produce likely reconstruction patterns in a much larger space.

While some of the considered state-of-the-art methods \cite{Du-AS-2019,Pham-CMIG-2019,ZENG-CBM-2018} presented results only for some cases on NAMIC, either 2D super-resolution on T1w images or 3D super-resolution on T2w images, we provide our results for all possible cases. We note that our CNN model surpasses Lanczos interpolation in each and every case. Furthermore, our model provides superior results than all the state-of-the-art methods \cite{Du-AS-2019,Pham-CMIG-2019,You-TMI-2019,ZENG-CBM-2018} considered in our evaluation on the NAMIC data set.

In addition to the quantitative results shown in Tables~\ref{tab_namic_results_2D} and \ref{tab_namic_results_3D}, we present qualitative results in Figure~\ref{fig_namic_results}. We selected 5 examples of 2D super-resolution results generated by Lanczos interpolation, by the GAN-based method of You et al.~\cite{You-TMI-2019} and by our CNN model. A close inspection reveals that our results are generally sharper than those of Lanczos interpolation and those of You et al.~\cite{You-TMI-2019}. As also confirmed by the SSIM, the PSNR and the IFC values presented in Tables~\ref{tab_namic_results_2D} and \ref{tab_namic_results_3D}, the images generated by our CNN are closer to the ground-truth images. At the scale factor of $2\times$ considered in Figure~\ref{fig_namic_results}, our CNN does not produce any patterns or artifacts that deviate from the ground-truth.

\subsection{Image Quality Assessment Results}

\begin{table}[!t]
\caption{Image quality assessment results collected from 6 doctors and 12 regular annotators, for the comparison between our CNN-based method versus Lanczos interpolation. For each upscaling factor, each annotator had to select an option for a number of 100 image pairs. To prevent cheating, the randomly picked locations (left or right) for the generated HR images were unknown to the annotators.}\label{tab_human}
\begin{center}
\begin{tabular}{|l|c|c|c|c|}
\hline 
Annotator ID  & \multicolumn{2}{|c|}{$2\times$} & \multicolumn{2}{|c|}{$4\times$} \\
\cline{2-5}   & Our CNN           & Lanczos           & Our CNN           & Lanczos \\
\hline   
\hline
Doctor {\#}1          & $100$ & $0$ & $100$ & $0$\\ 
\hline  
Doctor {\#}2          & $85$ & $15$ & $100$ & $0$\\   
\hline 
Doctor {\#}3          & $100$ & $0$ & $100$ & $0$\\  
\hline  
Doctor {\#}4          & $86$ & $14$ & $71$ & $29$\\ 
\hline  
Doctor {\#}5          & $100$ & $0$ & $68$ & $32$\\ 
\hline  
Doctor {\#}6          & $95$ & $5$ & $95$ & $5$\\  
\hline
Person {\#}1          & $100$ & $0$ & $99$ & $1$\\ 
\hline  
Person {\#}2          & $97$ & $3$ & $100$ & $0$\\ 
\hline  
Person {\#}3          & $98$ & $2$ & $100$ & $0$\\ 
\hline  
Person {\#}4          & $98$ & $2$ & $100$ & $0$\\   
\hline 
Person {\#}5          & $100$ & $0$ & $98$ & $2$\\  
\hline   
Person {\#}6          & $100$ & $0$ & $76$ & $24$\\ 
\hline  
Person {\#}7          & $100$ & $0$ & $93$ & $7$\\ 
\hline  
Person {\#}8          & $99$ & $1$ & $98$ & $2$\\ 
\hline  
Person {\#}9          & $100$ & $0$ & $98$ & $2$\\   
\hline 
Person {\#}10         & $100$ & $0$ & $98$ & $2$\\  
\hline 
Person {\#}11         & $98$ & $2$ & $99$ & $1$\\ 
\hline  
Person {\#}12         & $100$ & $0$ & $98$ & $2$\\ 
\hline  
Overall in \%         & $97.55 \%$ & $2.45  \%$ & $96.69\% $ & $3.31 \%$\\
\hline 
\end{tabular}
\end{center}
\end{table}

We provide the outcome of the subjective image quality assessment by human observers in Table~\ref{tab_human}. The study reveals that both doctors and regular annotators opted for our approach in favor of Lanczos interpolation at an overwhelming rate ($97.55\%$ at the $2\times$ scale factor and $96.69\%$ at the $4\times$ scale factor). For the $2\times$ scale factor, $10$ out of $18$ annotators preferred the output of our CNN in all the 100 presented cases. We note that doctors \#2 and \#4 opted for Lanczos interpolation in 15 and 14 cases (for the $2\times$ scale factor), respectively, which was not typical to the other annotators. Similarly, for the $4\times$ scale factor, there are 3 annotators (doctor \#4, doctor \#5 and person \#6) that seem to prefer Lanczos interpolation at a higher rate than the other annotators. After discussing with the doctors about their choices, we discovered that, in most cases, they prefer the sharper output of our CNN. However, the CNN seems to introduce some reconstruction patterns (learned from training data) that do not correspond exactly to the ground-truth. This phenomenon seems to be more prevalent at the $4\times$ scale factor, although the phenomenon is still rarely observed. This explains why doctors \#4 and \#5 preferred Lanczos interpolation in more cases than the other doctors, although the majority of their votes are still in favor of our CNN. When they opted for Lanczos interpolation, they considered that it is safer to consider its blurred and less informative output. In trying to find an explanation for these reconstruction patterns, we analyzed the output of the CNN without data augmentation. We observed such reconstruction patterns even when training data augmentation was removed, ruling out this hypothesis. Given a low-resolution input patch, the CNN finds the most likely high-resolution patch corresponding to the input. This likelihood is learned by the CNN when it is trying to minimize the loss over the entire training set. Although producing the most probable output works well in most cases, using a machine learning model, e.g. a CNN, is not the perfect solution. The explanation becomes clear if we consider that multiple HR patches can correspond to the same LR input patch and that choosing the most likely HR patch is not always the right answer. We thus conclude that our CNN suffers from the same problem as any other machine learning model. Furthermore, we stress out that the reconstruction patterns in question are plausible from a biological point of view, i.e. the doctors were able to spot them only by comparing the HR output with the ground-truth HR image. We note that these patterns should not be mistaken with artifacts that could be caused by underfitting or a poor architectural choice. Our CNN does not introduce such artifacts.

Based on our subjective image quality assessment, we concluded with the doctors that going beyond the $4\times$ scale factor, solely with a method based on algorithmic super-resolution, is neither safe (a CNN might introduce too many patterns far from the ground-truth) nor helpful (a standard interpolation method is not informative). However, the doctors agree that either super-resolution method is desirable in favor of the input low resolution images. Therefore, in order to reach the scale factor of $10\times$ desired by the doctors, we have to look in other directions in future work. A promising direction is to combine multiple inputs, e.g. by using CT and MRI scans of the same person or by using CT/MRI scans taken at different moments in time (before and after the contrast agent is introduced).

\section{Conclusion}
\label{sec_Conclusion}

In this paper, we have presented an approach based on fully convolutional neural networks for the super-resolution of CT/MRI scans. Our method is able to reliably upscale 3D CT/MRI image up to a scale factor of $4\times$. We have compared our approach with several baseline interpolation and state-of-the-art methods \cite{Du-AS-2019,Pham-CMIG-2019,You-TMI-2019,ZENG-CBM-2018}. The empirical results indicated that our approach provides superior results on both CH and NAMIC data sets. We have also conducted a subjective image quality assessment by human observers, showing that our method is significantly better than Lanczos interpolation. The subjective image quality assessment also revealed the limitations of a pure algorithmic approach. The doctors invited to take part in our study concluded that going to a scale factor higher than $4\times$ requires alternative solutions. In future work, we aim to continue our research by extending the proposed CNN method to multi-channel input. This will likely help us in achieving higher upscaling factors, e.g. $10\times$, required for the accurate diagnostics and treatment of cancer, an actively studied and extremely important research topic \cite{Popa-RRP-2014,Sardari-EBM-2010}.

\section*{Acknowledgment}
The authors would like to thank the reviewers for their helpful comments. In memory of Tatiana-Mihaela Ionescu, who offered her CT scans to facilitate our study, but died of cancer.


\ifCLASSOPTIONcaptionsoff
  \newpage
\fi



%

\bibliographystyle{IEEEtran}
\bibliography{IEEEabrv,references}

\EOD

\end{document}